\documentclass[10pt,twocolumn,letterpaper]{article}

\usepackage{cvpr}
\usepackage{times}
\usepackage{epsfig}
\usepackage{graphicx}
\usepackage{amsmath}
\usepackage{amssymb}
\usepackage{caption}
\usepackage{subcaption}
\usepackage[normalem]{ulem}

\usepackage[breaklinks=true,bookmarks=false]{hyperref}

\cvprfinalcopy 

\def\cvprPaperID{****} 

\begin{document}

\title{Technical Report: Image Captioning with Semantically Similar Images}

\author{Martin Kol\'{a}\v{r} \\
{\tt\small kolarmartin@fit.vutbr.cz} \and Michal Hradi\v{s},\\
{\tt\small ihradis@fit.vutbr.cz} \\
Faculty of Information Technology \\ Brno University of Technology\\
\and 
Pavel Zem\v{c}\'{i}k\\
{\tt\small zemcik@fit.vutbr.cz}
}

\maketitle

\begin{abstract}
This report presents our submission to the MS COCO Captioning Challenge 2015. The method uses Convolutional Neural Network activations as an embedding to find semantically similar images. From these images, the most typical caption is selected based on unigram frequencies. Although the method received low scores with automated evaluation metrics and in human assessed average correctness, it is competitive in the ratio of captions which pass the Turing test and which are assessed as better or equal to human captions.

\end{abstract}

\section{Introduction}


Image Captioning is a challenging problem which requires smart and careful combination of Computer Vision with Natural Language Processing. Our approach, presented in this report, yields results which outperform several state-of-art published methods while being significantly simpler.

In our approach, the last hidden layer of a Convolutional Neural Network is used as an embedding. For a given test image, we find the nearest training images, and retrieve their captions. In this body of captions, word counts are used to select the sentence which contains the most repeated terms, and this sentence is used to annotate the test image.

This simple method was entered into the Microsoft COCO~\cite{lin2014microsoft} 2015 captioning challenge, where it was evaluated by various widely used metrics, and assessed by human judges through Amazon Mechanical Turk. These results are presented and discussed here.


\section{Semantic Similarity Captioning}

In order to simplify our approach, we divided it into three steps: CNN embedding (\ref{cnn-embedding}), Finding similar images (\ref{similarity-metric}), and caption selection (\ref{sentence-selection}).

For a test image, a pre-trained image classification CNN is evaluated, and the last hidden layer is used as an embedding. In this embedding space, $n$ nearest training images are chosen. All the captions of these training images are then bagged together, unordered. Finally, one of these sentences is selected as the annotation of the test image.

\begin{table*}[t]
\begin{tabular}{ |l|c|c|c| }
  \hline
  assessment & our score & human & random\\
  \hline
  Ratio of captions that are evaluated as better or equal to human caption. & 0.194 & 0.638 & 0.007 \\
  Ratio of captions that pass the Turing Test. & 0.213 & 0.675 & 0.020 \\
  Average correctness of the captions on a scale 1-5 (incorrect - correct). & 3.079 & 4.836 & 1.084 \\
  Average amount of detail of the captions on a scale 1-5 (lack of details - very detailed). & 3.482 & 3.428 & 3.247 \\
  Ratio of captions that are similar to human description. & 0.154 & 0.352 & 0.013 \\
  \hline
\end{tabular}
\caption{Our scores on the MS COCO Captioning Challenge 2015}
\label{table-scores}
\end{table*}

\subsection{CNN Embedding}
\label{cnn-embedding}
We compute semantic image embedding using  the  Caffe reference network~\cite{jia2014caffe}, pretrained on ILSVRC~\cite{ILSVRC15} images.  Specifically, the embedding is provided by activations of the last hidden layer after the ReLU nonlinearity. The activations are a sparse vector of length $4096$ and were shown to be suitable for semantic content-based image search~\cite{Crowley14a} . This is evaluated on all training and test images.

\subsection{Finding Similar Images}
\label{similarity-metric}
\begin{figure}
\begin{subfigure}[b]{0.5\textwidth}
	\begin{center}
		\includegraphics[width=1\linewidth]{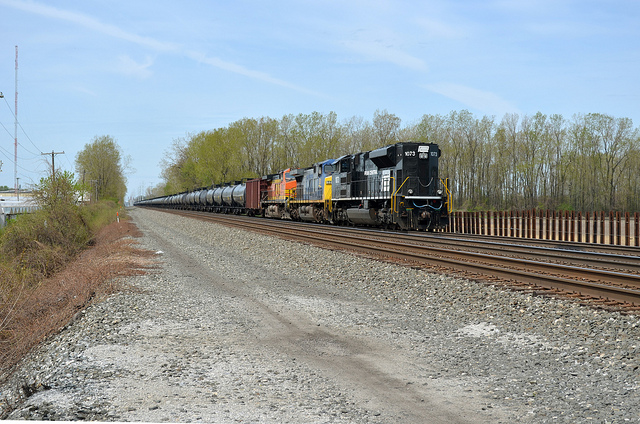}
		\caption{Query image}
	\end{center}
\end{subfigure}
\\

\begin{subfigure}[b]{0.5\textwidth}
        \begin{subfigure}[b]{0.3\textwidth}
		\includegraphics[width=1\linewidth]{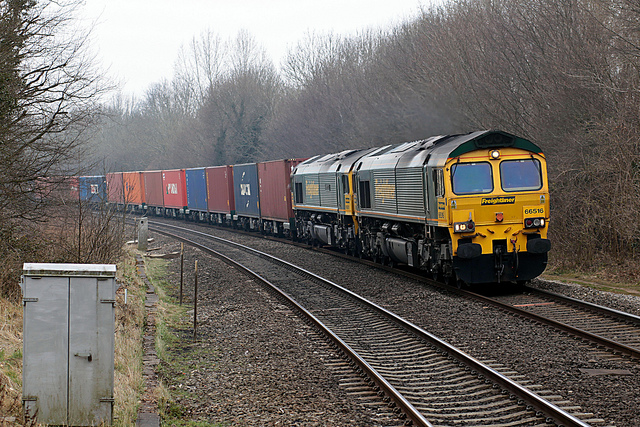}
                \label{fig:gull}
        \end{subfigure}
        ~ 
        \begin{subfigure}[b]{0.3\textwidth}
		\includegraphics[width=1\linewidth]{images/111981.jpg}
                \label{fig:tiger}
        \end{subfigure}
        ~ 
        \begin{subfigure}[b]{0.3\textwidth}
		\includegraphics[width=1\linewidth]{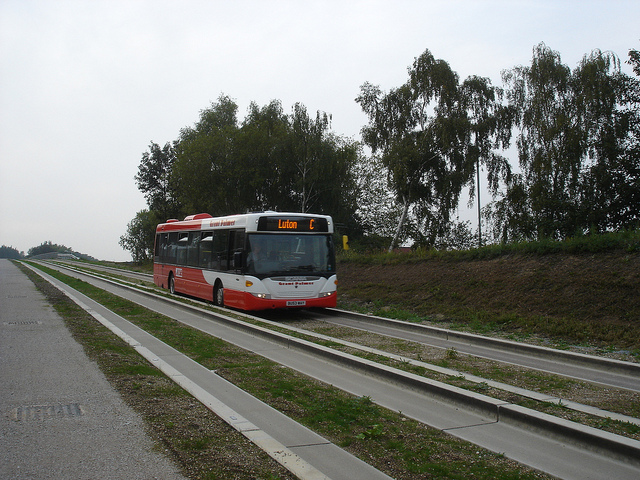}
                \label{fig:mouse}
        \end{subfigure}
        \begin{subfigure}[b]{0.3\textwidth}
		\includegraphics[width=1\linewidth]{images/68413.jpg}
                \label{fig:gull}
        \end{subfigure}
        ~ 
        \begin{subfigure}[b]{0.3\textwidth}
		\includegraphics[width=1\linewidth]{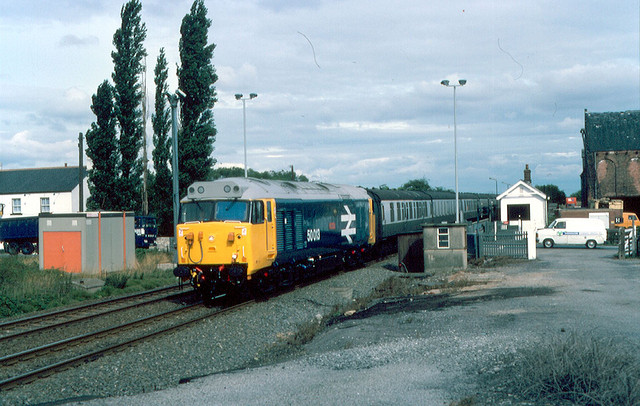}
                \label{fig:tiger}
        \end{subfigure}
        ~ 
        \begin{subfigure}[b]{0.3\textwidth}
		\includegraphics[width=1\linewidth]{images/49097.jpg}
                \label{fig:mouse}
        \end{subfigure}
        \begin{subfigure}[b]{0.6\textwidth}
		\includegraphics[width=1\linewidth]{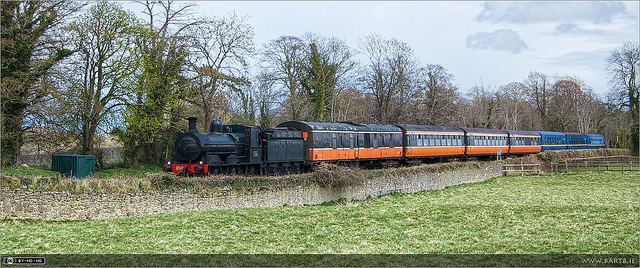}
                \label{fig:gull}
        \end{subfigure}
        ~ 
        \begin{subfigure}[b]{0.32\textwidth}
		\includegraphics[width=1\linewidth]{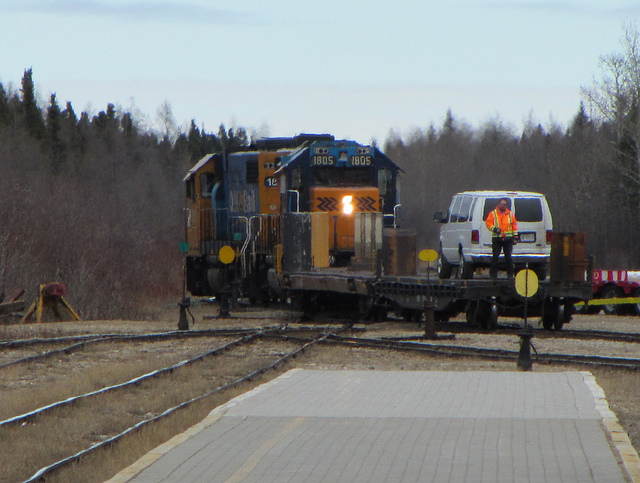}
                \label{fig:tiger}
        \end{subfigure}
        ~ 
        \begin{subfigure}[b]{0.26\textwidth}
		\includegraphics[width=1\linewidth]{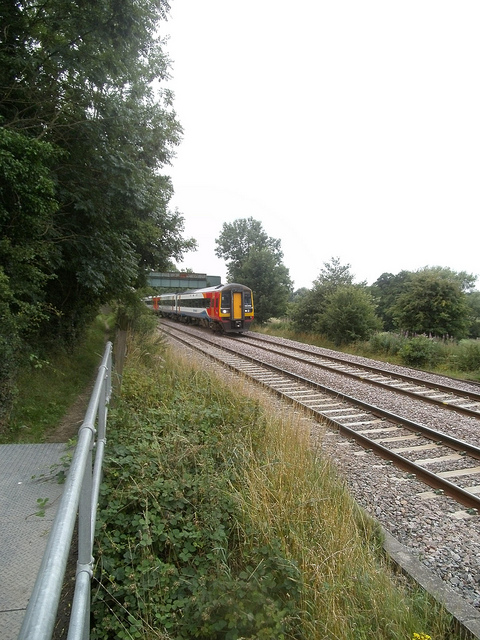}
                \label{fig:mouse}
        \end{subfigure}
        ~ 
        \begin{subfigure}[b]{0.5\textwidth}
		\includegraphics[width=1\linewidth]{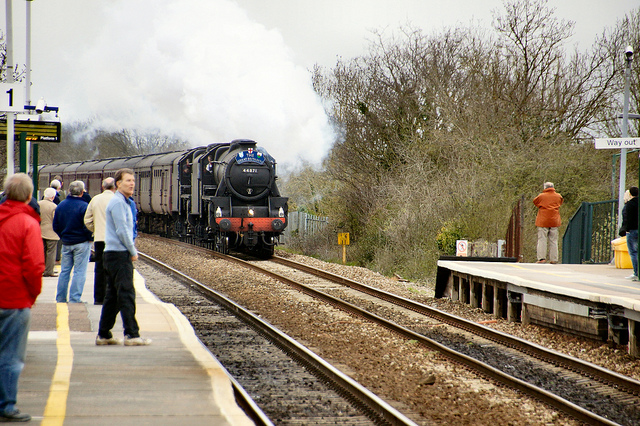}
                \label{fig:mouse}
        \end{subfigure}
        
        \caption{Retrieved images}\label{fig:animals}
\end{subfigure}
\caption{Example results of the intermediate steps}
\label{example}
\end{figure}

For a query image, we find $n$ nearest database images by cosine distance. See figure~\ref{example} for an example. $n$ is chosen manually, to fit the task, and we chose $n=10$ for the MS COCO 2015 challenge. We tested other distances (1-norm, 2-norm, $\infty$-norm, and ranking by linear SVM), but these did not outperform cosine distance.

\subsection{Caption Selection}
\label{sentence-selection}
All captions of the  $n$ most similar images are bagged to create a description corpus -- we ignore ranks of images. Since $n=10$ and $5$ captions are given for every database image, we obtain a corpus of $50$ candidate sentences, see figure~\ref{text}. From the candidate sentences, we select the most representative one by iteratively removing sentences which don't contain words which occur most frequently in the $50$ candidate sentences. This culling process starts from the most frequent word.
If a word is not present in any of the  remaining candidate sentences, it is skipped. The process ends when only one sentence remains. The $100$ most used words according to Google n-grams~\cite{michel2011quantitative} are ignored in the culling process. Table~\ref{table-words} shows words used in the example image.

\section{Results}
Figure~\ref{example} shows an example of images retrieved using the CNN embedding . As in this example, the retrieved images match semantically more than visually, as desired. Figure~\ref{text} shows all candidate sentences of the retrieved images, and Table~\ref{table-words} shows the words used to select the final caption.

In the MS COCO Captioning Challenge 2015, resulting captions were assessed by human judges according to five metrics. Table~\ref{table-scores} presents our score for each metric, along with the score for human and random annotation.


\section{Future Work}
Improvements can be made in a number of trivial ways: the dataset can be extended, the CNN can be trained or fine-tuned on the MS COCO images and categories, and caption selection can be weighted by the entropy of each word in the training corpus.

\section{Discussion}
Although improvements can be made on this method, we are of the opinion that this approach will only ever have limited potential. Rather than attempting to perform well on a given dataset, the goal in image captioning should be the creation of a model of scene understanding. Discriminative approaches such as this one can perform well in typical cases, but a generative approach is needed to understand and describe unlikely scenes.

We recommend to use this or similar method as a captioning baseline.

\newpage

{\small
\bibliographystyle{ieee}
\bibliography{egbib}
}

\begin{figure}
\center
\scriptsize
Guy stands near a train carrying gravel on its holding cars \\
A man standing next to train on a train track. \\
A train with multiple carts and a person working on it. \\
A man standing next to a train on train tracks. \\
A train attendant stands near a train pulling two filled cars. \\
A train hauling a van is crossing some railroad tracks. \\
Train passing a man on rural country road.  \\
A train with a man on the back of it with a vehicle in the background \\
A train car carrying a man and a white van \\
Much needed train track repairs are now in progress.  \\
A yellow bus carrying passengers riding along the road. \\
A yellow school bus driving down a street with a red car following behind it. \\
There is a bus driving down the street. \\
This yellow bus is driving down the street \\
A yellow bus traveling down a cobblestone road. \\
A train driving down the tracks near a platform. \\
A train that is sitting on a train track. \\
People on a platform watching a steam train pull in \\
A train traveling down tracks next to a loading platform. \\
A train driving toward a station where people are waiting. \\
A long train traveling through a tree covered countryside. \\
A multicolored freight train on a track amid greenery. \\
A train has many containers attached to it \\
A train traveling down the tracks through scenic scenery. \\
a long colorful passenger train going down a track by some trees \\
\footnotesize
\textbf{A train traveling down a train track next to trees.} \\
\scriptsize
A train on the tracks moving through bushes.  \\
an image of a train riding along the rail road track \\
A railroad train traveling down the train tracks \\
A train is approaching on a railroad track. \\
A train is riding down the tracks in the middle of some woods. \\
A very long large train going down a track. \\
A long train on a track next to another track. \\
A colorful train sits on the tracks on a foggy day. \\
A train with two engines pulling cars along the curve of a fall photo. \\
 A locomotive train engine is pulling cars along a railroad track. \\
A train with passenger cars on train tracks. \\
A train with an older locomotive drives through the country. \\
a train sitting on a track next to a bunch of trees \\
Steam train engine on the tracks in a field. \\
A train traveling past a building and two lights. \\
The train is passing by parked trucks in a lot and buildings.  \\
A train that is sitting on the tracks. \\
A passenger train that is traveling down some tracks. \\
a big train drives down a track through a city  \\
A bus moving down a road lane designated for buses. \\
A red and white bus traveling on a side road. \\
A bus moving fast along an interstate highway. \\
A red and white bus traveling the bus lane on a highway. \\
a public transit bus on an empty road 
\caption{Annotations from selected images, with the selected one in bold}
\label{text}
\end{figure}
\normalsize

\newpage

\begin{table}
\begin{center}

\scriptsize
\begin{tabular}{ |c|c| }
  \hline
  count & word\\
  \hline
  83 & \sout{a} \\
     47 & train \\
     21 & \sout{on} \\
     19 & \sout{the} \\
     15 & down \\
     14 & track \\
     13 & tracks \\
     11 & bus \\
     10 & traveling \\
     10 & \sout{is} \\
      7 & \sout{with} \\
      7 & \sout{to} \\
      7 & road \\
  \hline
\end{tabular}
\normalsize
\caption{Most frequent unigrams for the example image. Ignored words are crossed.}
\label{table-words}
\end{center}
\end{table}

\end{document}